\def\year{2022}\relax
\documentclass[letterpaper]{article} % DO NOT CHANGE THIS
\usepackage{aaai22}  % DO NOT CHANGE THIS
\usepackage{times}  % DO NOT CHANGE THIS
\usepackage{helvet}  % DO NOT CHANGE THIS
\usepackage{courier}  % DO NOT CHANGE THIS
\usepackage[hyphens]{url}  % DO NOT CHANGE THIS
\usepackage{graphicx} % DO NOT CHANGE THIS
\usepackage{natbib}  % DO NOT CHANGE THIS AND DO NOT ADD ANY OPTIONS TO IT
\usepackage{caption} % DO NOT CHANGE THIS AND DO NOT ADD ANY OPTIONS TO IT
\DeclareCaptionStyle{ruled}{labelfont=normalfont,labelsep=colon,strut=off} % DO NOT CHANGE THIS
\usepackage{algorithm}
\usepackage{algorithmic}

\usepackage{amsmath}
\usepackage{amssymb}
\usepackage{multirow}
\usepackage{booktabs}
\usepackage{newfloat}
\usepackage{listings}
\floatstyle{ruled}
\newfloat{listing}{tb}{lst}{}
\floatname{listing}{Listing}
\usepackage{balance}
\usepackage{xcolor}

\newcommand{\steven}[1]{\textcolor{blue}{#1}}

\title{Inferring Prototypes for Multi-Label Few-Shot Image Classification\\ with Word Vector Guided Attention}
\author{
    Kun Yan\textsuperscript{\rm 1},  Chenbin Zhang\textsuperscript{\rm 1}, Jun Hou\textsuperscript{\rm 2}, Ping Wang\textsuperscript{\rm 1}\thanks{corresponding author}, Zied Bouraoui\textsuperscript{\rm 3}, \\ Shoaib Jameel\textsuperscript{\rm 4}, Steven Schockaert\textsuperscript{\rm 5} 
}
\affiliations{
    \textsuperscript{\rm 1}Peking University, \textsuperscript{\rm 2}SenseTime Research, \textsuperscript{\rm 3}CRIL - University of Artois  \& CNRS, \\
    \textsuperscript{\rm 4}University of Essex, \textsuperscript{\rm 5}Cardiff University
}

\usepackage{bibentry}
\newif\ifappendix
\appendixtrue

\begin{document}

\maketitle

\begin{abstract}
%Multi-label few-shot image classification (ML-FSIC) is the task of learning to recognize a set of previously unseen categories of images from a small number of training examples. In this work, a cross-modal metric-based method is specifically designed for the ML-FSIC task. We first propose a global semantic constraint module to construct a joint feature space which aligns features from different modalities. Then, we develop a category-specific feature generation module to extract the visual prototype for each category from images which have the corresponding category. As one image may have multiple categories, a cross-model multi-head attention mechanism is introduced, which can pay more attention to the category-specific local features given the query of text features of the corresponding category names. Our method can extract category-specific prototype for unseen categories based on a small number of training examples without finetuning model parameters, which shows the strong generalization ability. Extensive experiments on COCO and PASCAL VOC datasets demonstrate that our approach can consistently achieve the state-of-the-art performance in multi-label few-shot image classification task. Our codes and dataset settings will make public soon.
Multi-label few-shot image classification (ML-FSIC) is the task of assigning descriptive labels to previously unseen images, based on a small number of training examples. A key feature of the multi-label setting is that images often have multiple labels, which typically refer to different regions of the image. When estimating prototypes, in a metric-based setting, it is thus important to determine which regions are relevant for which labels, but the limited amount of training data makes this highly challenging. As a solution, in this paper, we propose to use word embeddings as a form of prior knowledge about the meaning of the labels. In particular, visual prototypes are obtained by aggregating the local feature maps of the support images, using an attention mechanism that relies on the label embeddings. 
%The pre-computed label embeddings are thus essentially used to determine which areas of which support images are most likely to be relevant for that label. 
As an important advantage, our model can infer prototypes for unseen labels without the need for fine-tuning any model parameters, which demonstrates its strong generalization abilities. Experiments on COCO and PASCAL VOC furthermore show that our model substantially improves the current state-of-the-art.
\end{abstract}

\section{Introduction}
%Multi-label image classification~(ML-IC) is a fundamental yet practical task in computer vision, as real-word images generally contain multiple objects. As a result of this, ML-IC has attracted a great quantity of researches~\cite{cnn-rnn,chen2019multi,recurrently-dis,orderless}. However, the methods mentioned above depend on a large amount of training data. As a contrast, the human visual system can learn to recognize new classes with only a few instances. This observation fuels us on designing an algorithm to solve a more generative task: multi-label few-shot image classification (ML-FSIC), which aims to learn to recognize a set of new categories in an image based on a small number of training examples.

Multi-label image classification (ML-IC) has received considerable attention in recent years \cite{cnn-rnn,chen2019multi,recurrently-dis,orderless}. The aim of this task is to assign descriptive labels to images, where each image is typically associated with multiple labels. Standard approaches for this task often focus on modelling label dependencies, e.g.\ taking advantage of the fact that the presence of one label makes the presence of another label more (or less) likely. In the few-shot setting, however, we only have a small number of images available for training, possibly only a single image for some labels. Clearly, relying on label co-occurrence statistics is not feasible in such a setting.

%Compared to ML-IC, ML-FSIC is more challenging, as there are only a few available training examples, which is far from enough to train ML-IC methods. We find directly apply ML-IC methods to ML-FSIC task can get poor performance, even some methods~\cite{chen2019multi,you2020cross} based on graph convolutional networks~\cite{gcn} can not be used, where they need obtain statistical information among labels from training examples. Because in ML-FSIC, with only a few training examples, in which the statistical relationship of different labels is not credible. 

%Recently, the problem of few-shot image classification~(FSIC) has received much attention, many kinds of approaches are proposed, like meta-learning-based~\cite{MAML,Meta-SGD} and metric-based methods~\cite{protonet,few-shot-gnn,feat}, etc. However, FSIC methods aim at the single-class problem, which predict one label for each image. In contrast, there is little work on multi-label few-shot image classification. 

The problem of few-shot image classification (FSIC), i.e.\ image classification with limited training data in the single-label setting, has also received considerable attention. However, standard approaches for this task are not suitable for the multi-label setting. For instance, so-called metric-based approaches learn a prototype for each image category, and then assign images to the category whose prototype is closest to the image in some sense. These prototypes are typically obtained by averaging a representation of the training images. In the seminal ProtoNet model \cite{protonet}, for instance, prototypes are simply defined as the average of the global feature maps of the available training examples. This strategy crucially relies on the assumption that most of the image is somehow relevant to its category. In the multi-label setting, however, such an assumption is highly questionable, given that different labels tend to refer to different parts of the image. For instance, given an image depicting a car and a bike, using a representation of the entire image to obtain a prototype for bike would be misleading.

Our aim in this paper is to introduce a metric-based model for multi-label few-shot image classification (ML-FSIC). Given the aforementioned concerns, we need a strategy that is based on local image features, allowing us to focus on those parts of the training images that are most likely to be relevant. However, as we may only have a single training example for some labels, we cannot implement such a strategy without some kind of prior knowledge about the meaning of the labels. We will rely on word vectors \cite{glove} for this purpose. Some previous works for the single-label setting have already relied on word vectors for inferring prototypes directly~\cite{am3, aligning}, but as the resulting prototypes are inevitably noisy, such strategies are most useful in combination with prototypes that are derived from visual features. Therefore, taking a different approach, in this paper we only use word vectors to identify which regions of the training images are most likely to be relevant for a given label. As an example to explain the intuition of how word vectors can be useful for this purpose, assume that we have a number of labels that refer to animals. These labels will have similar word vectors, which tells the model that the predictive visual features for these different labels are likely to be similar. Now suppose we have an image which is labelled with \emph{cat}. Based on training data for other labels, the model will select areas that are likely to contain an animal (although it would not necessarily be able to distinguish between cats and closely related animals). Note that word embeddings are thus used as prior knowledge about the similarity of different labels. %, rather than for predicting visual prototypes. 
An important practical advantage of our method is that we can apply the model to previously unseen labels, without the need for any fine-tuning of the model's parameters on the novel label set. To the best of our knowledge, our model is also the first end-to-end method for ML-FSIC.

As another contribution, we propose a number of changes to the evaluation methodology for ML-FSIC systems. The most important change is concerned with how support sets are sampled, as part of an episode based strategy. 
The standard $N$-way $K$-shot framework for evaluating FSIC systems is based on the idea that exactly $K$ training images are available for each category of interest. 
While earlier work in ML-FSIC has aimed to mimic this $N$-way $K$-shot framework as closely as possible, we found this to have significant drawbacks when images can have multiple labels. 
We also propose some changes related to how the query set is sampled and the choice of evaluation metrics. Finally, we propose a new ML-FSIC dataset based on PASCAL VOC~\cite{pascal}, which is a standard ML-IC dataset that we adapt for the few-shot setting.

\smallskip
\noindent To summarize, the contributions of this work are as follows:
\begin{itemize}
\item To the best of our knowledge, we propose the first metric-based method for multi-label few-shot image classification. %After training the model in the base set, we can directly infer images without any finetuning in the novel set, which contains categories that are disjoint with ones in base set. We are also the first end-to-end method for ML-FSIC. 
\item We propose a model that uses word vectors as prior knowledge about labels in a novel way. %Rather than using word vectors to initialise Graph Neural Networks~\cite{kggr}, as has been proposed for ML-IC, or to infer prototypes directly~\cite{am3}, as has been proposed for FSIC, we use word vectors to guide an attention mechanism.
%\item We propose a cross-modality weights based loss to align the semantic information of visual and text features in a joint space, and further develop a cross-modal multi-head attention algorithm to extract the category-specific visual features guided by text features. 
\item We propose a number of methodological changes, to obtain an evaluation framework that is better suited for ML-FSIC than existing strategies. 
\item We introduce a new benchmark based on PASCAL VOC.
%\item We re-stantardize the settings of multi-label few-shot learning based on the several problem observations, like sampling strategy, data split and so forth. We also propose a new ML-FSIC dataset based on PASCAL VOC. Our method can achieve consistently the state-of-the-art performance on both COCO and PASCAL VOC datasets.  
\end{itemize}

% \begin{figure*}[t]
% \centering
% \includegraphics[width=0.95\textwidth]{figure1} % Reduce the figure size so that it is slightly narrower than the column.
% \caption{Overview of our approach. There are mainly two components: global semantic constraint~(GSC) module and category-specific feature generation~(CFG) module. GSC is developed for constructing a joint space, where features in different modality of the same category have similar semantic information, via a cross-modal weights based loss~(CMW-Loss). CFG is proposed to extract category-specific visual features from local features of all images, which have the corresponding category, in the support set. The feature generated by CFG is the prototype for the corresponding category, which can be the classification weights.}
% \label{fig1}
% \end{figure*}

% \begin{figure*}[t]
% \centering
% \includegraphics[width=0.95\textwidth]{LaTeX/figure2.pdf} % Reduce the figure size so that it is slightly narrower than the column.
% \caption{\yankun{Overview of our approach to get visual prototypes for labels. The prototypes are extracted from local features of all images which have the corresponding labels in the support set. The cross-modality multi-head attention encoder is developed for extracting prototypes guided by pre-trained label embeddings.}}
% \label{fig2}
% \end{figure*}

\section{Related Work}
In this section, we review the related work on \emph{multi-label image classification}, \emph{few-shot image classification}, and the combined area of \emph{multi-label few-shot image classification}.

%following two main research streams: multi-label image classification and image classification in few-shot scenario. The latter one contains single-label and multi-label few-shot image classification. 

\subsection{Multi-label Image Classification}
Early solutions for ML-IC simply learned a binary classifier for each label \cite{tsoumakas2007multi}. More recently, various methods have been proposed to improve on this basic strategy by exploiting label dependencies in some way. For instance, the CNN-RNN architecture \cite{cnn-rnn} learns a joint embedding space for representing both images and labels, which is used to predict image-label relevance. Because the labels are represented as vectors, semantic dependencies are implicitly taken into account. To avoid the need for a predefined label order, as in RNN based architectures, \citet{orderless} proposed minimal loss alignment~(MLA) and predicted label alignment~(PLA) to dynamically order the ground truth labels with the predicted label sequence. Some studies~\cite{wang2020multi,chen2019multi,you2020cross} also exploit graph convolutional networks~(GCN)~\cite{gcn} to model label dependencies more explicitly. Recently, \citet{multi-label-trans} used transformers to better exploit the complex dependencies among visual features and labels. However, the above methods require a large amount of training data, and can thus not be directly applied in the few-shot setting.

As mentioned in the introduction, attention mechanisms play an important role in ML-IC, to associate labels with specific image regions. For instance, \citet{recurrently-dis} proposed a spatial transformer layer to locate attentional regions in convolutional feature maps, and applied an LSTM sub-network to sequentially predict labels from the resulting regions. \citet{spatial-reg} proposed the spatial regularization network to generate attention maps for all labels and captures the underlying relations via learnable convolutions. These existing strategies again require sufficient training data, and are thus not suitable for the few-shot setting.

%\subsection{Image Classification in Few-shot Scenario}
\subsection{Few-Shot Image Classification}
Different strategies for single-label few-shot image classification have already been proposed, with  metric-based~\cite{relationnet,edge_labeling,few-shot-gnn,feat} and meta-learning based~\cite{optimization-as-model,MAML,Meta-SGD} methods being the most prominent. %, although some other directions have also been explored, such as hallucination based~\cite{low-shot-visual,low-shot-data,MetaGAN} methods. 
Meta-learning based methods use a meta-learner to learn to adapt model parameters to new categories in the few-shot regime. Our method is more closely related to metric-based methods, which aim to learn a generalizable visual embedding space in which different image categories are spatially separated.
%where data from different categories can be distinguished with distance metrics. 
ProtoNet~\cite{protonet} generates a visual prototype for each class by simply averaging the embeddings of the support images in this embedding space. The category of a query image is then determined by its Euclidean distance to these prototypes. Instead of using Euclidean distance, the Relation Network~\cite{relationnet} learns to model the distance between query and support images. 
%\steven{Graph convolutional networks have also been used for this purpose}\nb{Is this what you meant? If so, do you have a reference to such a paper? Or did you mean that the Relation Network relies on GCNs? In that case, can you clarify what kind of dependencies are being modelled?}.  
Other notable models include FEAT~\cite{feat}, which uses a transformer to contextualize the image features relative to the support set and PSST~\cite{chen2021pareto}, which introduced a self-supervised learning strategy. 

While most models rely on global features, methods exploiting local features have also been proposed \cite{revisit,local-mine}, but these methods are designed for single-label classification. For instance, \citet{revisit} calculate the similarity between all local features of the query image and all local features of the support images. As such, there is no attempt to focus on particular regions of the support images. %and use the top-$k$ similarity as the final score. and summed up the top\_k similarity scores as the final score of the corresponding category.\yankun{local-mine} selected representative local features according to the change of classification loss. However, the change of classification loss can not distinguished different labels. Both of them assign the same label to all local features of an image. In other words, they can not handle singe-label images. } 
The use of word vectors has also been considered, for instance for estimating visual prototypes \cite{am3,aligning}. However, due to the inevitably noisy nature of the predicted prototypes, such methods are best used in combination with prototypes obtained from visual features. %In particular, they are not suitable for the setting we consider in this paper. 
Word vectors have also been used in margin-based models, to adaptively set margins based on the semantic similarity between categories~\cite{traml}, and for grouping visual features into facets \cite{multi-facet}.

%However, single-label few-shot image classification methods can only predict one label for each image, they can not solve multi-label problems. 

\subsection{Multi-Label Few-Shot Image Classification}
The ML-FSIC problem has only received limited attention.
%In contrast, multi-label few-shot learning problem has attracted much less attention. 
LaSO~\cite{laso} was the first model that was designed to address this problem. It relies on a data augmentation strategy which generates synthesized feature vectors via label-set operations. KGGR~\cite{kggr} uses a GCN to take label dependencies into account, where labels are modelled as nodes and two nodes are connected if the corresponding labels tend to co-occur. The strength of these label dependencies is normally estimated from co-occurrence statistics, but for labels with limited training data, dependency strength is instead estimated based on GloVe word vectors \cite{glove}.
%proposed a knowledge-guided graph routing framework, which uses GloVe word vectors \cite{glove} to extract semantic vector for new categories \nb{This sounds conceptually quite similar to what we're doing in this paper; can you add some explanation of how their method differs from what we're doing?} \yankun{The KGGR method is based on GCN. GCN has edges and nodes. The nodes are features of categories. The edge weights are usually co-occur possibility of two categories.  The edge weights are usually calculated in the training set. However, for novel set, only a few-shot examples are available. They can't calculate the statistic information for novel categories. Therefore, they use GloVe vectors to compute the similarity of two categories instead.} , rather than computing the statistical cooccurrence from training examples. 
\citet{li2021compositional} proposed an ML-FSIC method which learns compositional embeddings based on weak supervision. Due to its use of weak supervision, this method is not directly comparable with our method. LaSO and KGGR, on the other hand, focus on the same supervised setting that we consider in this paper. However, our approach has the advantage that model parameters do not need to be updated to predict previously unseen labels, which makes our model easier to use than LaSO and KGGR. Moreover, unlike these two existing methods, our model can be trained end-to-end.

%We mainly compare our method with the first two\nb{say why}\yankun{The method~\cite{li2021compositional}is based on weak-supervision. Not supervised method like us or LaSO or KGGR.} , however, both of them need finetune model parameters in novel set, which cost much time. Besides, both of them need two-stage training strategy, they are not end-to-end. On the contrary, after being trained in base set, our method can directly predict labels for novel set. Our model can be trained end-to-end. 

\section{Problem Setting}\label{secSetting}
We consider the following multi-label few-shot image classification~(ML-FSIC) setting. We are given a set of base labels $\mathcal{C}_{\textit{base}}$ and a set of novel labels $\mathcal{C}_{\textit{novel}}$, where $\mathcal{C}_{\textit{base}}\cap\mathcal{C}_{\textit{novel}} = \emptyset$. We also have two sets of labelled images: $\mathcal{E}_{\textit{base}}$, containing images with labels from $\mathcal{C}_{\textit{base}}$, and  $\mathcal{E}_{\textit{novel}}$, containing images with labels from $\mathcal{C}_{\textit{novel}}$, where $\mathcal{E}_{\textit{base}}\cap\mathcal{E}_{\textit{novel}} = \emptyset$. The images from $\mathcal{E}_{\textit{base}}$ are used for training the model, while those in $\mathcal{E}_{\textit{novel}}$ are used for testing. The goal of ML-FSIC is to obtain a model that performs well for the labels in $\mathcal{C}_{\textit{novel}}$, when given only a few examples of images that have these labels. 
Models are trained and evaluated using so-called episodes. Each training episode involves a support set and a query set. The support set corresponds to the examples that are available for learning to predict the labels in $\mathcal{C}_{\textit{base}}$, while the query set is used to assess how well the system has accomplished this goal.

To construct the support set of a given episode, for every label in $\mathcal{C}_{\textit{base}}$, we sample one image from $\mathcal{E}_{\textit{base}}$ which has that label. These images are sampled without replacement, meaning that the total number of images in the support set is given by $|\mathcal{C}_{\textit{base}}|$. The query set is sampled in the same way, except that we sample a larger number of images per label. Testing episodes are constructed similarly, but with labels from $\mathcal{C}_{\textit{novel}}$ and images from $\mathcal{E}_{\textit{novel}}$ instead.

%$N$ classes from $\mathcal{C}_{\textit{novel}}$ are sampled, and examples in support set are made available for training, while the remaining images are then used to construct query set for testing. After sampling many episodes, the average result of all episodes can reflect the performance of the model.

Note that our strategy for sampling episodes differs from the standard strategy from FSIC. In particular, FSIC models are usually evaluated using episodes that contain a sub-sample of $N$ classes, where the support set contains exactly $K$ examples of each class. In ML-FSIC, this strategy is difficult to adopt, since each image may have multiple labels. The idea of setting $N=|\mathcal{C}_{\textit{base}}|$ during training and $N=|\mathcal{C}_{\textit{novel}}|$ during testing conforms to the strategy that was used by \citet{laso} and \citet{kggr}. However, \citet{laso} fix the number of training examples per label as $K$, with $K\in\{1,5\}$,
  %In previous few-shot image classification~(FSIC), it usually follows a $N$-way $K$-shot setting~($N$ and $K$ usually equal to 1 or 5), which means in each episode, $N$ classes are sampled, $K$ labelled examples from each class are sampled to construct support set. However, in ML-FSIC, as one image may have several labels, it is hard to follows the same $N$-way $K$-shot setting in FSIC. Therefore, we make the $N$ setting the same as in \cite{laso} and \cite{kggr}, which $N$ is defined by dataset. In other words, $N$ is equal to the number of classes of $\mathcal{C}_{\textit{novel}}$ in dataset. As for $K$, in \cite{laso}, it is still strictly controlled as 1 or 5 just like in FSIC. 
  which has two important shortcomings. First, there are typically only few combinations of images that can be selected to construct support sets, while adhering to the requirement that each label has to occur $K$ times (even when somewhat relaxing this requirement). This means that only a few episodes can be sampled, which makes it harder to train the model, and makes the test results less stable, as they are averaged over a small number of test episodes.
  %Even in \cite{laso}, it pointed out the amount of labels per category could exceed $K$. 
Second, the total number of images in the support set can vary substantially. For example, if one image contains all labels, then we may have a support set that only contains that one image when $K = 1$. 
  
  %As defining the $K$ as a fixed number is not feasible in ML-FSIC, to make the support set construction more generable and easier, we sample one training image for each label without repetition, which can make sure each label is contained by at least one image, and the number of training images of different episode is the same \nb{For training you take one image from all the labels in $C_{\textit{base}}$, while for testing you take one image from all the labels in $C_{\textit{test}}$, right? Do you ensure that the images that are used during training are excluded from the test set? Because some images can presumably have some labels from $C_{\textit{base}}$ and some labels from $C_{\textit{test}}$}.

\begin{figure}[t]
\centering
\includegraphics[width=0.95\columnwidth]{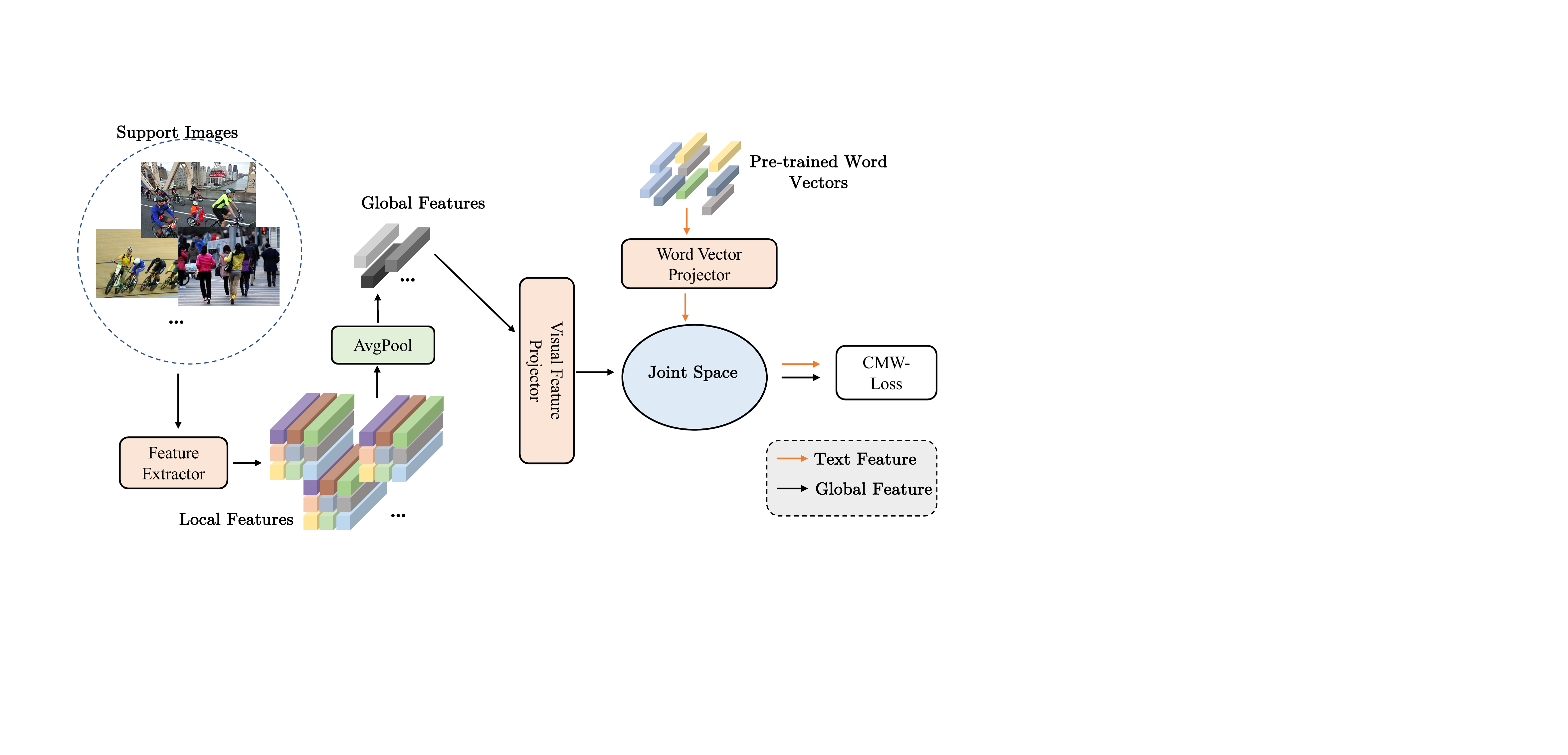} 
\caption{A joint embedding space is learned in which both labels and images are represented.}
%\yankun{CMW-Loss is proposed to constrain the joint space, where features in different modality of the same category have similar semantic information.}
\label{fig1}
\end{figure}

\begin{figure}[t]
\centering
\includegraphics[width=0.95\columnwidth]{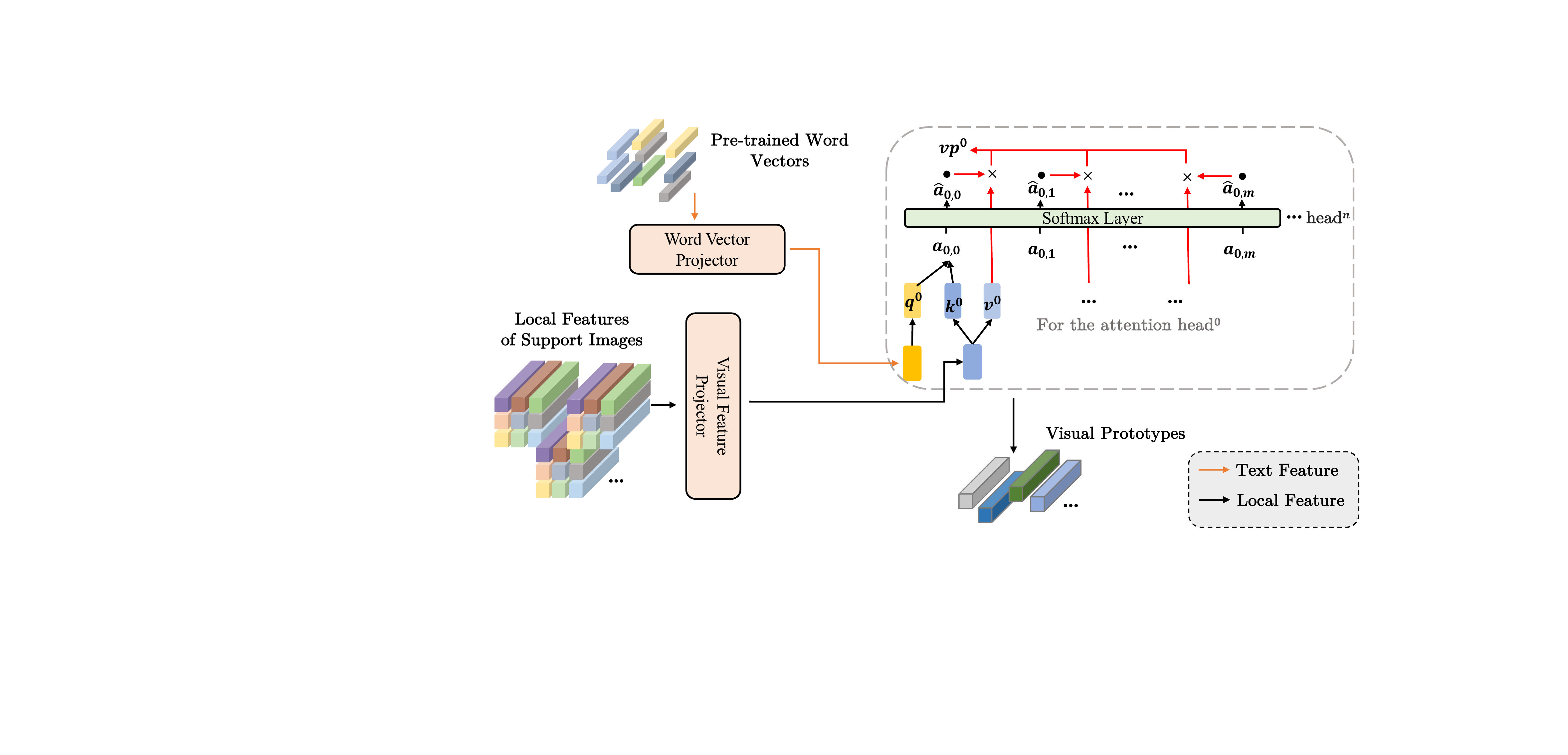} % Reduce the figure size so that it is slightly narrower than the column.
\caption{An attention mechanism is used to compute label prototypes from the local features of the relevant images from the support set. The embedding of the considered label plays the role of \emph{query} in this attention mechanism.}
%\yankun{Overview of our approach to get visual prototypes for labels. The prototypes are extracted from local features of all images which have the corresponding labels in the support set. The cross-modality multi-head attention encoder is developed for extracting prototypes guided by pre-trained label embeddings.}}
\label{fig2}
\end{figure}

\begin{figure}[t]
\centering
\includegraphics[width=0.95\columnwidth]{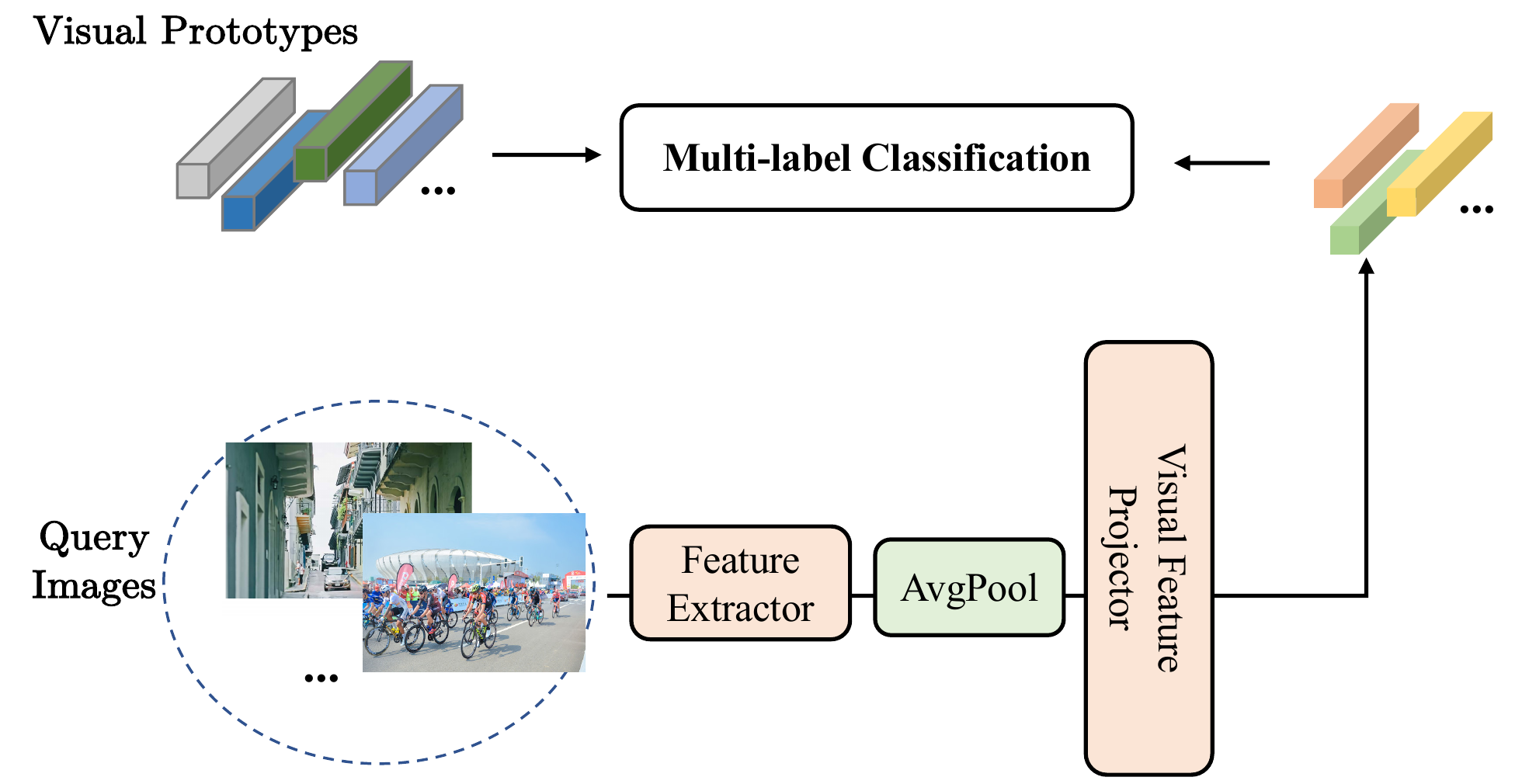} 
\caption{Query images are classified based on their distance to the visual prototypes.}
%\yankun{Visual prototypes are as classification weights to classify query images.}}
\label{fig3}
\end{figure}

\section{Method}
Our model consists of two main components. The first component is aimed at jointly representing label embeddings and visual features in the same vector space. Essentially, this component aims to predict visual prototypes from the label embeddings. Since such prototypes are noisy, we do not use them directly for making the final label predictions. This component is merely used to learn a joint representation of visual features and labels; it is illustrated in Figure \ref{fig1}. The second component is aimed at computing the final prototypes, by aggregating the local features of the corresponding support images based on an attention mechanism, which relies on the label representations that are obtained by the first component. This second component is illustrated in Figure \ref{fig2}. Finally, to classify a query image, we project it to the joint embedding space and then compare it with the learned prototypes, as illustrated in Figure \ref{fig3}. We now describe these different steps in more detail.

%The overview of our proposed architecture is shown in Fig.~\ref{fig1}. There are two steps in our method. (i) We first construct a joint space to align different modal features, where features in different modality of the same category have similar semantic information. In this way, we can guide visual model by text features. (ii) We need to construct classification weights for each category, which we call the prototypes. As in multi-label scenario, one image may have multiply labels, we need to extract the category-specific features from images. To this end, we develop two components: global semantic constraint~(GSC) module and category-specific feature generation~(CFG) module for the two steps respectively.

%\subsection{Global Semantic Constraint Module}
\subsection{Joint Embedding of Visual Features and Labels}\label{jointLearning}
Given an input image $I$, we first use a feature extractor to obtain its local feature map $\mathbf{f}_{\textit{loc}}^{I} \in \mathbb{R}^{n \times h \times w}$, where $n$ is the number of channels, $h$ is the height and $w$ is the width. In this paper, we use a fully convolutional network such as ResNet~\cite{residual} for this purpose. 
%\begin{equation}
	%\mathbf{f_l}^{I} = f_{\textit{cnn}}(I),
%\end{equation}
%where $f_{cnn}(\cdot)$ is the feature extractor, and it is implemented by a fully convolutional network, like Resnet~\cite{residual}. 
The global visual feature vector $\mathbf{f}_{\textit{glo}}^{I}$ for image $I$ is obtained by feeding the local feature map through an adaptive average pooling layer:
\begin{equation}
	\mathbf{f}_{\textit{glo}}^{I} = \textit{AdaptiveAvgPool}(\mathbf{f}_{\textit{loc}}^{I})
\end{equation}
We use pre-trained word embeddings to represent the labels in $\mathcal{C}_{\textit{base}}\cup \mathcal{C}_{\textit{novel}}$. Let us write $\mathbf{w}_c$ for the word vector representing label $c$, and let $d_w$ be the dimensionality of the word vectors.
%We resort to word embeddings to apply text information. For each category, we first extract the word embedding vector using the pre-trained language model~(like GloVe~\cite{glove}) according to the name of the category, formulated as 
%\begin{equation}
	%\mathbf{w}_{c} = f_{g}(n_{c})
%\end{equation}
%where $\mathbf{w}_{c}$, $n_{c}$ are the word embeddings and name of the category $c$.
With the aim of representing images and labels in the same vector space, we learn two linear transformations:
\begin{align*}
\widehat{\mathbf{f}_{\textit{glo}}^I} &= \mathbf{A}_{\textit{visual}}\, \mathbf{f}_{\textit{glo}}^I  &
\widehat{\mathbf{w}_c} &= \mathbf{A}_{\textit{text}}\, \mathbf{w}_{c}
\end{align*}
where $\mathbf{A}_{\textit{visual}}\in \mathbb{R}^{d_j \times n}$ is used to project the global feature vector for $I$ onto a space of $d_j$ dimensions. Similarly,  $\mathbf{A}_{\textit{text}}\in \mathbb{R}^{d_j \times d_w}$ is used to project the $d_w$-dimensional embedding of a label $c$ onto the same $d_j$-dimensional space. 
%
%As $\mathbf{f_g}$ and $\mathbf{w}$ are in different dimensionality and modality, we first use two independent projectors to project both $\mathbf{f_g}$ and $\mathbf{w}$ in a joint space, where the dimensionality of  different modal features is the same. We use $\widehat{\mathbf{f_g}}$ and $\widehat{\mathbf{w}}$ to present $\mathbf{f_g}$ and $\mathbf{w}$ in the joint space respectively.
%
%The aim of the projections $\mathbf{A}_{\textit{visual}}$ and $\mathbf{A}_{\textit{text}}$ is to allow us to represent images and labels in the same space.
To ensure that the resulting image vectors $\widehat{\mathbf{f}_{\textit{glo}}^I}$ and label representations $\widehat{\mathbf{w}_c}$ are semantically compatible, we use the following loss:
\begin{align*}
	\mathcal{L}_{\textit{cmw}} & = \sum_{I\in \mathcal{S}}\sum_{i=1}^{|\mathcal{C}|} y_{i}^I \cdot \log \sigma(s_{i}^I) + (1 - y_{i}^I \cdot \log(1 - \sigma(s_{i}^I)))
\end{align*}
where $\mathcal{S}$ represents the set of images from the support set of the current training episode, $\mathcal{C}=\{c_1,...,c_{|\mathcal{C}|}\}$ is the set of labels, $\sigma(\cdot)$ is the sigmoid function and $y_i^I$ represents the ground truth, i.e.\ $y_i^I=1$ if image $I$ has label $c_i$ and $y_i^I=0$ otherwise. Finally, we have 
\begin{equation}
	s_{i}^I = \lambda\cos(\widehat{\mathbf{f}_{\textit{glo}}}^{I}, \widehat{\mathbf{w}_{c_i}}) 
\label{eq4}
\end{equation}
The scalar $\lambda$ is a hyper-parameter to address the fact that the cosine is bounded between -1 and 1.
Since the aim of the loss 	$\mathcal{L}_{\textit{cmw}}$ is to align two different modalities (word vectors and visual features), we refer to it as the Cross-Modality Weights Loss (CMW-loss).

%To further constrain this joint feature space to have the attribute that features in different modality of the same category have similar semantic information. We propose a cross-modality weights based loss~(CMW-Loss). Concretely, we use the word embeddings as the classification weights to classify all labeled images. Given a global feature of an image~($\widehat{\mathbf{f_g}}^{I}$), we first calculate the cosine similarity between $\widehat{\mathbf{f_g}}^{I}$ and word embeddings of all categories. The similarity between $\widehat{\mathbf{f_g}}^{I}$ and the word embedding of category $c$ is formulated as

%Based on the similarity score from Eq.\ref{eq4}, we use the binary-cross-entropy loss to update the model to align different modal features, formulated as

%where $n$ is the number of all classes, $\sigma(\cdot)$ is the sigmoid function. Considering the value of the raw $s_{c} \in [-1, 1]$, to make the model easier to train, we multiply $s_{c}$ by a constant $\gamma$ as $s_{c} = s_{c} \cdot \gamma$. we find $\gamma = 20$  is more than enough, therefore, we set the $\gamma = 20$ in the later experiments.   

%\subsection{Category-specific Feature Generation Module}
\subsection{Constructing Attention Based Prototypes}\label{consAtten}
%To extract the category-specific visual features from an image which contains multiple categories, we propose a cross-modal multi-head attention mechanism, which treat the word embeddings as query, treat local features as key and value, can generate the category-specific visual features for the corresponding category. 
We now explain how the label prototypes are constructed. Consider a label $c$ and suppose this label has been assigned to $m$ images from the support set. We can obtain a total of $l= h \cdot w \cdot m$ local feature vectors from these $m$ images. Let us write these local feature vectors as $\mathbf{u}_1,...,\mathbf{u}_l$. We first map these feature vectors to the joint embedding space: %that was learned by the CMW-loss:
\begin{align*}
\widehat{\mathbf{u}_i} = \mathbf{A}_{\textit{visual}}\, \mathbf{u}_i
\end{align*}
The prototype of $c$ will be obtained from these local feature vectors using an attention mechanism that is inspired by \citet{attention}, where the label embedding $\widehat{\mathbf{w}_c}$ is used as the query component. Specifically, we have
\begin{align*}
\mathbf{q}^j_c &= \mathbf{Q}_j \widehat{\mathbf{w}_c}\\
\mathbf{k}^j_i &= \mathbf{K}_j \widehat{\mathbf{u}_i}\\
\mathbf{v}^j_i &= \mathbf{V}_j \widehat{\mathbf{u}_i}\\
(\mu^j_1,...,\mu^j_l) &= \textit{softmax}\left(\frac{\mathbf{q}^j_c\cdot \mathbf{k}^j_1}{\sqrt{d_a}},...,\frac{\mathbf{q}^j_c\cdot \mathbf{k}^j_l}{\sqrt{d_a}}\right)\\
\mathbf{p}_c^j &= \sum_i \mu_i^j \mathbf{v}^j_i
\end{align*}
where $d_a$ is the dimensionality of the vectors $\mathbf{q}^j_c$, $\mathbf{k}^j_i$, and $\mathbf{v}^j_i$.
The vector $\mathbf{p}_c^j$ represents the contribution of the j\textsuperscript{th} attention head to the prototype of label $c$. We use a total of $n_a$ attention heads. The final prototype of label $c$ is given by:
\begin{align}
\mathbf{p}_c = \textit{mlp}(\mathbf{p}_c^1 \oplus ... \oplus \mathbf{p}_c^{n_a})\label{eqPrototypeComputation}
\end{align}
where we write $\oplus$ for vector concatenation and \textit{mlp} consists of two fully connected feedforward layers with GeLU activation and dropout. Since the prototype $\mathbf{p}_c$ should be $d_j$ dimensional, %the vectors $\mathbf{q}^j_c$, $\mathbf{k}^j_i$ and $\mathbf{v}^j_i$ are chosen to be $\frac{d_j}{n_a}$-dimensional, i.e.\ 
we have $d_a=\frac{d_j}{n_a}$.

To train the attention mechanism, we use the following loss function, which we refer to as the Query Loss:
\begin{align*}
	\mathcal{L}_{\textit{query}} & = \sum_{I\in\mathcal{Q}}\sum_{i=1}^{|\mathcal{C}|} y_{i}^I \cdot \log \sigma(q_{i}^I) + (1 - y_{i}^I) \cdot \log(1 - \sigma(q_{i}^I))
\end{align*}
where $\mathcal{Q}$ represents the set of images from the query set of the current training episode. As before, $\mathcal{C}$ is the set of labels and $y_i^I$ represents the ground truth. The predictions $q_i$ are obtained as follows:
\begin{align}
	q_{i}^I = \lambda\cos(\widehat{\mathbf{f}_{\textit{glo}}}^{I}, \mathbf{p}_{c_{i}}) 
\label{q_p}
\end{align}
with $\lambda$ the same scalar as in \eqref{eq4}. Note that there are two key differences between $\mathcal{L}_{\textit{cmw}}$ and $\mathcal{L}_{\textit{query}}$: (i) $\mathcal{L}_{\textit{cmw}}$ is trained using the support images while $\mathcal{L}_{\textit{query}}$ is trained using the query images; and (ii) prototypes in $\mathcal{L}_{\textit{cmw}}$ are estimated from word vectors while prototypes in $\mathcal{L}_{\textit{query}}$ are those obtained by aggregating local visual features. %using the attention mechanism.

\begin{table*}[t]
\centering
\footnotesize
\begin{tabular}{l cccc  cccc}
\toprule
 & \multicolumn{4}{c }{\textbf{Micro}} & \multicolumn{4}{c}{\textbf{Macro}}\\
\cmidrule(l){2-5} \cmidrule(l){6-9}
& \textbf{Prec} & \textbf{Recall} & \textbf{F1} & \textbf{AP} & \textbf{Prec} & \textbf{Recall} & \textbf{F1} & \textbf{AP} \\
\midrule
ResNet-50   & 9.96 & 17.62 & 12.51 & 11.31    & 10.25 & 17.69 & 13.27 & 19.67 \\ 
ResNet-101  & 9.81 & 17.65 & 12.04 & 10.22    & 9.48  & 17.86 & 12.18 & 18.87  \\ 
ViT         & 9.63 & 12.96 & 11.02 & 10.07    & 9.22  & 14.16 & 11.13 & 16.94 \\ 
ResNet-50 + ViT    & 8.21 & 13.51 & 10.21 & 10.13    & 9.15 & 14.72 & 11.17 & 16.71  \\ 
PLA    & 13.29 & 55.44 & 20.89 & 21.33    & 13.24 & 54.79 & 20.96 & 30.61  \\ 
PLA~(GloVe)    & 14.12 & 55.83 & 22.15 & 23.02    & 13.51 & 53.21 & 20.37 & 30.13  \\ 
LaSO            & 12.31 & 19.77 & 15.09 & 16.83    & 13.64 & 20.03 & 16.11 & 25.41 \\ 
MAML   & 15.42 & 53.21 & 23.11 & 25.42    & 15.63 & 51.63 & 23.82 & 35.30 \\
\midrule
Ours   & 49.72 & 26.60 & \textbf{34.21} & \textbf{35.30}   & 34.50 & 25.07 & \textbf{28.91} & \textbf{42.84}\\ 
\bottomrule
\end{tabular}
\caption{Experimental results for COCO.}
\label{coco}
\vspace{-2mm}
\end{table*}

\subsection{Model Training and Evaluation}
The model is trained by repeatedly sampling training episodes from $\mathcal{C}_{\textit{base}}$, as explained in Section \ref{secSetting}. Given a training episode with support set $\mathcal{S}$ and query set $\mathcal{Q}$, the model parameters are updated using the following loss:
\begin{align*}
    \mathcal{L}_{\textit{all}} = \mathcal{L}_{\textit{cmw}} + \gamma \mathcal{L}_{\textit{query}}
\end{align*}
with $\gamma$ a hyperparameter to control the relative importance of both components. These components are defined as above, where $\mathcal{C}$ represents the the set $\mathcal{C}_{\textit{base}}$ during training.

%where $\mathcal{L}_{\textit{cmw}}$ is calculated based on the support set, while $\mathcal{L}_{\textit{query}}$ is from the query set\nb{What is the advantage of doing it like this? Wouldn't you get better results when using both the support and query set for both components of the loss function?}\yankun{I also tried to use query images for CMW-loss, but it did not give much help. Later, in order to finetune the model during test, we can only use support set images for CMW-loss, because labels only are available in support set, I finally decide only use support images for CMW-loss. As the reason why I did not use prototypes to classify support images, this is because I want to unify training and test phases. During test, only query images need to be classified by prototypes, support set images are only used to construct prototypes. Better results are usually achieved by unifying the training and testing phases. }. And $\gamma$ is used to control the distribution of $\mathcal{L}_{\textit{query}}$, we set $\gamma$ to 1 during training.. 
%}

After the model has been trained, it can be evaluated on test episodes as follows. For each episode, we first construct the prototypes using the support set, as in \eqref{eqPrototypeComputation}. Note that we can do this without fine-tuning any model parameters. For each query image $I$, the probability that it has label $c_i$ is computed as $\sigma(q_i^I)$ with $q_i^I$ as defined in \eqref{q_p}. 
% \nbyankun{I also tried before, the finetuning of our method only contains CMW-loss, as labels are only availabel on the support set. However,the improvement is limited, only increasing 3 percentage points on macro AP  and 0.4 percentage points on micro AP. So I didn't show them.}

%We also introduce the episode-based training scheme in ML-FSIC, which inspired by \cite{match-net} in FSIC. Concretely, during training the model in the base set, we constantly sample episodes following the form of episode in the test phase. Each time, $N$ classes are sampled~($\mathcal{C}_{\textit{base}}$), and we continue sample one image for each class from the base set to construct support set, while still sample $M$ images for each class to construct query set. We use the global features of images in support set and word embeddings to learn jointed feature space via CMW-Loss, and use the local features of images in support set and word embeddings to generate the prototypes for each class sampled in the current episode. The labels of query images are determined by the similarity between its feature and prototypes of categories. 

\begin{table*}[t]
\centering
\footnotesize
\begin{tabular}{l cccc  cccc}
\toprule
 & \multicolumn{4}{c }{\textbf{Micro}} & \multicolumn{4}{c}{\textbf{Macro}}\\
\cmidrule(l){2-5} \cmidrule(l){6-9}
& \textbf{Prec} & \textbf{Recall} & \textbf{F1} & \textbf{AP} & \textbf{Prec} & \textbf{Recall} & \textbf{F1} & \textbf{AP} \\
\midrule
ResNet-50   & 14.11 & 39.60 & 20.77 & 16.57     & 12.96 & 39.69 & 21.92 & 27.24 \\ 
ResNet-101  & 16.58 & 42.85 & 23.86 & 19.23    & 15.13  & 42.66 & 22.17 & 30.11  \\ 
ViT  	& 13.68 & 25.52 & 17.76 & 16.69     & 13.16 & 25.61 & 17.15 & 27.72 \\ 
ResNet-50 + ViT  & 11.96 & 16.19 & 13.70 & 18.71    & 11.14 & 16.24 & 12.86 & 28.76 \\ 
PLA    & 23.45 & 86.44 & 36.81 & 41.98    & 24.24 & 87.50 & 37.83 & 47.29 \\ 
PLA~(GloVe)    & 22.67 & 88.03 & 35.84 & 41.31    & 23.16 & 88.64 & 36.55 & 46.54  \\  
LaSO   & 18.71 & 48.48 & 27.02 & 22.12    & 17.11 & 48.12 & 25.20 & 32.50 \\
\midrule
Ours   & 26.78 & 83.97 & \textbf{40.19} & \textbf{46.28}  & 29.64 & 85.44 & \textbf{43.35} & \textbf{53.26} \\  
\bottomrule
\end{tabular}
\caption{Experimental results for PASCAL VOC.}
\label{voc}
\vspace{-3mm}
\end{table*}

\section{Experiments}
\subsection{Experimental Setup}
\subsubsection{Datasets}
We have conducted experiments on two datasets. First, we used COCO~\cite{lin2014microsoft}. This dataset was already used by \citet{laso}, who proposed a split into 64 training labels and 16 test labels. However, as they did not include a validation split, we split their 64 training labels into 12 labels for validation (\textit{cow}, \textit{dining table}, \textit{zebra}, \textit{sandwich}, \textit{bear}, \textit{toaster}, \textit{person}, \textit{laptop}, \textit{bed}, \textit{teddy bear}, \textit{baseball bat}, \textit{skis}) and 52 labels for training, while keeping the same 16 labels for testing. %(\textit{bicycle}, \textit{boat}, \textit{stop sign}, \textit{bird}, \textit{backpack}, \textit{frisbee}, \textit{snowboard}, \textit{surfboard}, \textit{cup}, \textit{fork}, \textit{spoon}, \textit{broccoli}, \textit{chair}, \textit{keyboard}, \textit{microwave}, \textit{vase}). 
We include images from the COCO 2014 training and validation sets.
% \nbyankun{COCO2014 is a public popular image detection dataset, it has train and validation sets. The test set is given without labels, because they open a website for the submission of detectors to calculate their performance. Therefore, train and validation sets mains the all labeled images in COCO 2014. There is also COCO2017, but both of them have the same images. The difference is the split of train and validation images. I claim COCO2014, because LaSO used COCO2014}
%\nb{I'm still somewhat confused. Is there a COCO dataset, to which you add the COCO2014 images, or is COCO2014 the only dataset that you're using?} \yankun{COCO2014 itself is a dataset, we only use COCO2014. }
The images which do not contain any of the test and validation labels are used as the training set. Similarly, the validation set only contains images that do not contain any training or test labels.
Second, we propose a new ML-FSIC dataset based on PASCAL VOC \cite{pascal}, which has 20 labels. To use as many images as possible, we select the following six labels for the novel set $\mathcal{C}_{\textit{novel}}$: \textit{dog}, \textit{sofa}, \textit{cat}, \textit{potted plant}, \textit{tv monitor}, \textit{sheep}. The following six labels were selected for the validation split: \textit{boat}, \textit{cow}, \textit{train}, \textit{aeroplane}, \textit{bus}, \textit{bird}. The remaining eight labels are used for training. We use the images from the VOC 2007 training, validation and test splits, as well as the VOC 2012 training and validation splits (noting that the labels of the VOC 2012 test split are not publicly available). We again ensure that the training images do not contain any labels from the validation and test splits, and the validation images do not contain any test labels.

%\subsubsection{Training and Test Setting}
\subsubsection{Methodology}
\label{train+test}
Every model is trained for 200 epochs, with the first 10 epochs used as warm-up. We used the Adam optimizer with an initial learning rate of 0.001. In contrast to our model, existing methods require a fine-tuning step during the test phase. We set the number of epochs for this fine-tuning step to 40. During the test phase, we sample 200 test episodes.  Different from \citet{laso}, in addition to macro-AP, we also report micro-AP. Moreover, following usual practice in ML-IC, we also report the (macro and micro) precision, recall and F1 metrics, where we assume that a label is predicted as positive if its estimated probability is greater than 0.5~\cite{spatial-reg}.
 %Macro metrics are evaluated by averaging per-class predictions, while micro metrics are overall measures which counts true predictions for all images of all  Among these metrics, macro/micro AP are the most important metrics that can provide a more comprehensive evaluation. 
%Methods except ours need a two-stage process to train the model. In the first stage, we train the feature extractor using training examples of the base categories. In the second stage, we fix the parameters of the extractor, and train the classifier using the support images of the novel categories.  In contrast, our model only need to be trained by training examples of base categories, and the trained model can be applied directly on novel categories based on the corresponding support images. 

\subsubsection{Implementation Details}
We use \emph{ResNet-50} and \emph{ResNet-101}~\cite{residual} as %convolutional network for 
feature extractors. 
%Two projectors in GSC are implemented as fully-connected layers\nb{Are these linear projections, like it is explained at the moment, or do you use a non-linearity? In any case, this should be explained in Section 4 rather than here. What we should report here is simply how the hyperparameters were set (like the dimensionality)}\yankun{no non-linearity, just a fully-connected layer} \nb{What is the dimensionality of the joint space?}.\yankun{512}
The dimensionality of the joint embedding space was set as $d_j=512$. 
%The hidden size~($d_{k}$) of the cross-modal multi-head attention in the CFG is set as 512\nb{I'm confused. So the dimensionality of the attention model is not determined by the dimensionality of the joint embedding space and the number of attention heads?}\yankun{The dimensionality is only determined by the joint embedding space. This 512 because the dimensionality in joint space is 512. The number of attention heads does't affect the dimensionality. The more heads, the smaller the dimensionality of the output of each head. This can make sure the finally concatenation vector (prototype) is 512-dimensional.  }. 
%The number of attention heads $n_a$ was set to 8. 
When sampling episodes, we sample one image for each label to construct the support set and four images for each label to construct the query set. As word embeddings, we considered 300-dimensional GloVe \cite{glove} vectors\footnote{Specifically, we used vectors that were trained from Wikipedia 2014 and Gigaword 5, which we obtained from the GloVe project page, at \url{https://nlp.stanford.edu/projects/glove}.}. We similarly used standard pre-trained fastText and word2vec embeddings from an online repository\footnote{https://developer.syn.co.in/tutorial/bot/oscova/pretrained-vectors.html}. Based on the validation split, for both datasets, the number of attention heads was set to 8, the 300-dimensional GloVe vectors were selected as the word embedding model, and the hyper-parameter $\gamma$ was set to 1.

%episodes from training set, we sample 5 images for each label, while 1 of them is used to construct support set, the left 4 images are for constructing query set. 
%The word embeddings are GloVes with 300 dimensionality unless otherwise specified. \steven{We fix the hyper-parameter $\lambda$ in \eqref{eq4} as 20.}

%\subsection{Comparisons with Other Approaches}
\subsection{Baselines}
%We conducted experiments on both COCO and PASCAL VOC datasets. 
We compare with LaSO~\cite{laso} and KGGR~\cite{kggr}, which were designed for the ML-FSIC setting. To put the results in context, we also compare with some methods that were designed for ML-IC. First, following LaSO~\cite{laso}, we attach a standard multi-label classifier to a number of different feature extractors: \emph{ResNet-50}, \emph{ResNet-101}, \emph{ViT-Base}~\cite{vit} and \emph{ResNet-50} + \emph{ViT-Base}. %Note that the \emph{ResNet} models are fully convolutional networks, while the \emph{ViT} models are transformer based networks. 
We also report results for the recent state-of-the-art CNN-RNN based method PLA~\cite{orderless}, which to the best of our knowledge has not previously been evaluated in the ML-FSIC setting. 
With the exception of KGGR, we used the original source code of the different baselines to produce the results. 
Since we did not have access to the KGGR source code, we only compare our method against the published results from the original paper, which followed the experimental setting from LaSO. %For each method except the KGGR, we report all metrics mentioned in Experimental Setup. 
For PLA, LaSO and our method, we used \emph{ResNet-50} as the feature extractor in the main experiments. However, since the reported results of KGGR are based on ResNet-101 and GoogleNet-v3, we used the latter models as feature extractors in the comparison with KGGR.

\subsection{Experimental Results}
The experimental results for COCO are shown in Table~\ref{coco}, showing that our proposed method outperforms the other methods by a substantial margin. %We can achieve the state-of-the-art in both micro/macro F1 and AP, which are the most two importance metrics in ML-FSIC. 
It is evident that large models are prone to overfitting, noting that in terms of model size, we have: \emph{ResNet-50} + \emph{ViT-Base} \textgreater \emph{ViT-Base} \textgreater \emph{ResNet-101} \textgreater \emph{ResNet-50}. This is not unexpected given the small number of labeled examples in ML-FSIC. LaSO can improve the ResNet-50 baseline because of its data augmentation strategy. Somewhat surprisingly, PLA performs better than LaSO, which shows that its LSTM component is able to model label dependencies in a meaningful way. PLA also needs word embeddings, but in the original model these embeddings are learned from the training data itself. For comparison, we also report results for a variant where these word vectors are initialised using the 300-dimensional GloVe vectors instead; the results are shown as PLA~(GloVe). As can be seen, using pre-trained word vectors does not lead to a meaningful improvement over the original PLA model. 
Finally, we also add results for MAML~\cite{MAML}, which is a popular meta-learning method for FSIC\steven{, }but which was not specifically designed for the ML-FSIC.
To compare with KGGR, we use the 1-shot and 5-shot settings proposed by LaSO. 
As shown in Table~\ref{previous}, for ResNet-101, 
in the 1-shot setting we achieved a macro-AP of 55.73, compared to 52.3 for KGGR. In the 5-shot setting, we achieved a macro-AP of 68.12, compared to 63.5 for KGGR.
Moreover our method can also consistently outperform both KGGR and LaSO in combination with GoogleNet-v3~\cite{szegedy2016rethinking}.
Experimental results for PASCAL VOC are shown in Table~\ref{voc}. We can again see that our proposed method achieves the best results. %Different from the results for COCO, in this case the ViT model outperforms ResNet-50 in terms of AP. 

\begin{table}[b]
\centering
\footnotesize
\begin{tabular}{lcc}
\toprule
\textbf{Method} & \textbf{1-shot} & \textbf{5-shot}\\
\midrule
LaSO~(GoogleNet-v3)    & 45.3 & 58.1 \\
KGGR~(GoogleNet-v3)    & 49.4 & 61.0 \\
KGGR~(ResNet-101)      & 52.3 & 63.5 \\
\midrule
Ours~(GoogleNet-v3)                & 53.41 & 65.07 \\
Ours~(ResNet-101)                  & \textbf{55.73} & \textbf{68.12} \\
\bottomrule
\end{tabular}
\caption{Experimental results following the common data and evaluation setting from LaSO \cite{laso}.}
\label{previous}
\end{table}

% \begin{figure*}[t]
% \centering
% \includegraphics[width=0.95\textwidth]{LaTeX/fig_vis.pdf} % Reduce the figure size so that it is slightly narrower than the column.
% \caption{The visualization of the attention weights of local features when generating prototypes for categories. Images of first two lines are from novel set on the COCO dataset, while the image of the last line is from validation set on the COCO dataset.}
% \label{fig_vis}
% \end{figure*}

\begin{figure}[t]
\centering
\includegraphics[width=0.95\columnwidth]{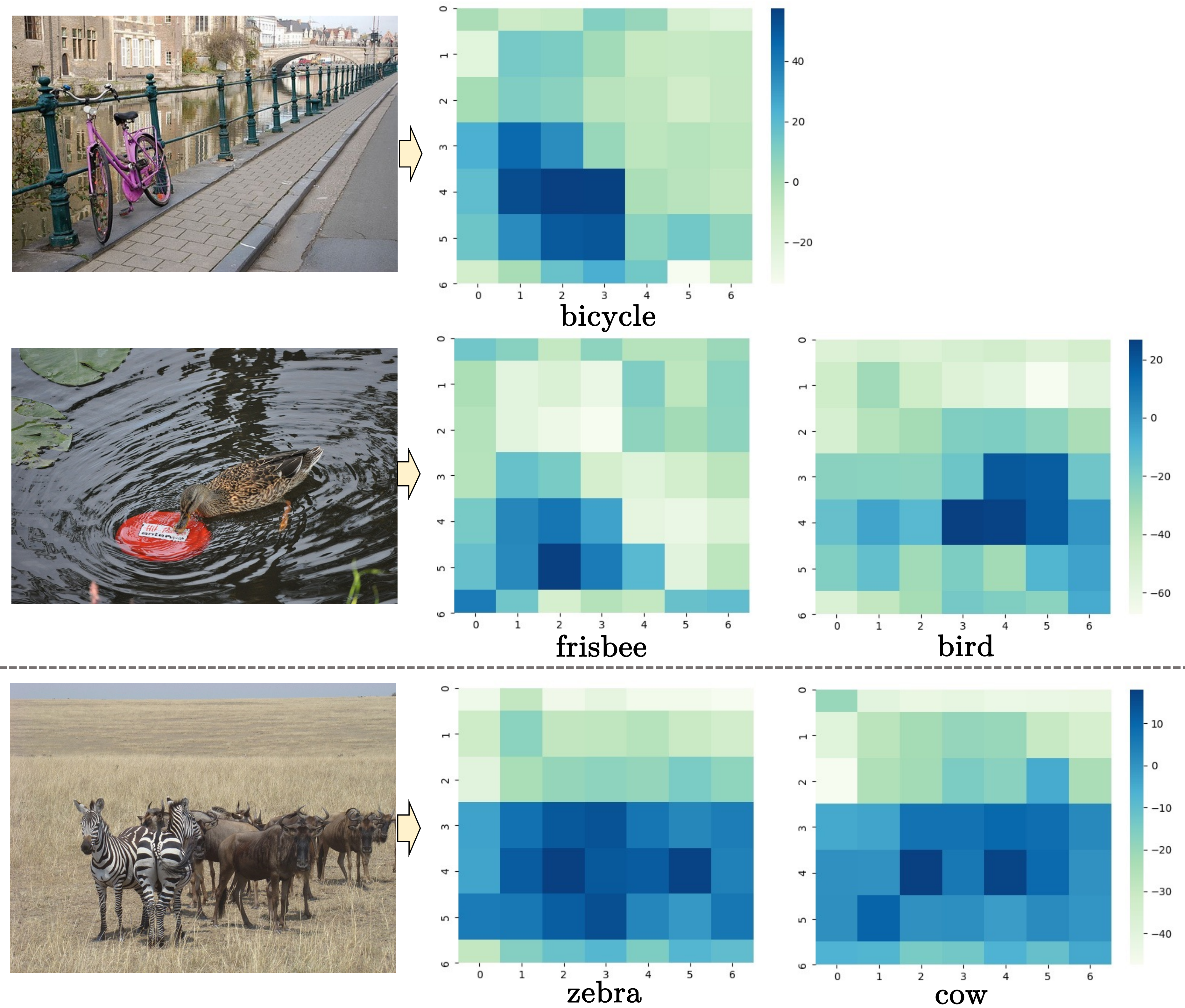} % Reduce the figure size so that it is slightly narrower than the column.
\caption{Visualization of the attention weights of local features for construction label prototypes. The first two examples are taken from the COCO test set, while last example is taken from the COCO validation set.}
\label{fig_vis}
\end{figure}

\begin{table}[t]
\centering
\footnotesize
\begin{tabular}{lcc}
\toprule
\textbf{Method} & \textbf{Macro-AP} & \textbf{Micro-AP}\\
\midrule
Without CMW-loss            & 36.17 & 26.44 \\
Simple attention            & 37.53 & 28.56 \\
% \yankun{Replace local features with global features}      & \yankun{35.82} & \yankun{27.39} \\
Attention with global features      & 35.82 & 27.39 \\
Simple att.\ with loc.\ features                     & 31.21 & 21.07 \\
Low-rank bilinear pooling  & 33.41 & 24.69 \\
Full model        & 42.84 & 35.30 \\
\bottomrule
%\noalign{\smallskip}
\end{tabular}
% \caption{Results for two variations of the proposed model.}
\caption{Results for different variations of the proposed model.}
\label{table1}
\end{table}

\begin{table}[t]
\centering
\footnotesize
\begin{tabular}{lcc}
\toprule
\textbf{Word Embeddings} & \textbf{Macro-AP} & \textbf{Micro-AP}\\
\midrule
FastText          & 35.67 & 24.72 \\
Skip-Gram         & 38.67 & 29.47 \\
GloVe-50          & 27.99 & 17.73 \\
GloVe-100         & 33.48 & 21.52 \\
GloVe-200         & 39.13 & 28.86 \\
GloVe-300         & 42.84 & 35.30 \\
BERT              & 36.46   & 28.33\\
\bottomrule
%\noalign{\smallskip}
\end{tabular}
\caption{Results for different word embeddings.}
\label{table3}
\vspace{-2mm}
\end{table}

\begin{table}[t]
\centering
\footnotesize
\begin{tabular}{lcc}
\toprule
\textbf{Method} & \textbf{Macro-AP} & \textbf{Micro-AP}\\
\midrule
Without Fine-tuning        & 42.84 & 35.30 \\
With Fine-tuning            & 45.27 & 35.70 \\
\bottomrule
%\noalign{\smallskip}
\end{tabular}
\caption{Impact of fine-tuning during the test phase.}
\label{table5}
\vspace{-2mm}
\end{table}

\subsection{Ablation Study}
Here we analyze the importance of the main components of our model. Some additional ablation analysis is provided in the appendix.
All experiments in this section are conducted on the COCO dataset with the \emph{ResNet-50} feature extractor. 

\subsubsection{Importance of the CMW-Loss}
To analyze the effect of the CMW-loss, in Table \ref{table1} we report the results that are obtained when this component of the loss function is removed. In particular, for this variant, we only use $\mathcal{L}_{\textit{query}}$ for training (using both the images from the support set and the query set in this case). %\nb{But in this case, the support images are not used at all for training the model, right? Would that not be the main reason for the decreased performance? }\yankun{The support images are used. Because we need support images to construct prototypes. The prototypes are used to classify query images. Therefore, support images also affect the back propagation process of the model} .
Note that the alignment between the word vectors and visual features then has to be learned indirectly, together with the parameters of the overall model. As can be seen, removing the CMW-loss results in substantially lower macro-AP and micro-AP scores.  %a decrease of 6.67 percentage points in macro-AP and 8.86 percentage points in micro-AP, which clearly demonstrates the importance of the CMW-loss.

%Global Semantic Constraint Module~(GSC), we only retain CFG to train the model. However, as we need the dimensionality of visual features and word embeddings the same apply CFG, we still keep the projectors in the GSC. In other word, during training phase, we remove the CMW-Loss which can constrain the joint space. 

%The results are shown in Table~\ref{table1}, removing the CMW-Loss results in a decrease of 6.67 percentage points in macro-AP and 8.86 percentage points in micro-AP. This demonstrate the important effect of CMW-Loss, which can align different modal features. Constrained by CMW-Loss, the constructed joint space has the property that features in different modality of the same category have similar semantic information. This property can support CFG to select category-specific visual features guided by word embeddings.

\subsubsection{Importance of the Attention Mechanism}
To evaluate the effect of the proposed attention mechanism, in Table \ref{table1} we also report results for a variant that relies on a simpler mechanism for generating prototypes (shown as \emph{Simple attention}). In particular, in this variant, we generate the prototype of a label $c$ by taking a weighted average of the global features of the support images that have that label. The weights are obtained by computing the cosine similarity between the vectors $\widehat{\mathbf{w}_c}$ and the feature vectors $\widehat{\mathbf{f}^I_{\textit{glo}}}$, multiplying these cosine similarities with the scalar $\lambda$, and feeding the resulting values to a softmax layer.
%Finally, we take a weighted average of global features of selected support images to generate the prototype for $c$.
As shown in Table~\ref{table1}, in this simplified setting, there is a drop of 5.31 and 6.74 percentage points in macro-AP and micro-AP respectively. We furthermore report results of a variant where the (full) attention mechanism uses global features instead of local features (\textit{Attention with global features}). We also report results when local features are used instead of global features in the simplified attention mechanism (\textit{Simplified attention with local features}).
Finally, we experimented with a variant that learns visual prototypes in the same way as in KGGR, where the weights of local features are generated by a low-rank bilinear pooling method followed by a fully-connected layer with softmax activation (\textit{Low-rank bilinear pooling}).
%\yankun{Besides, another two variants are also tested. (i) We replace local features in the attention mechanism by global features instead. (ii) We simplify the attention mechanism. Concretely, the prototype of a label is generated by taking a weighted average of local features of support images that have that label. The weights are obtained by computing the cosine similarity between the vectors $\widehat{\mathbf{w}_c}$ and the feature vectors $\widehat{\mathbf{f}^I_{\textit{loc}}}$, multiplying these cosine similarities with the scalar $\lambda$, and feeding the resulting values to a softmax layer. 
The results in Table~\ref{table1} show that all of these variants result in lower performance.

%We analyze that is because global features may have information of multiple categories (if an image has multiple labels). Besides, there are also background factors. The model is difficult to be robust to these noisy information with only a few training examples. With CFG, the model can select the category-specific local features guided by word embeddings via a cross-modal multi-head attention mechanism. The final prototypes are pure compared to global features. Therefore, we can get better results.

\subsubsection{Fine-Tuning}
Although our method can be used without fine-tuning during the test phase, it is possible to fine-tune the parameters with the CMW-Loss. %Specifically, for each test episode, we fine-tune the joint space by aligning support image features and word vectors based on CMW-Loss. 
As shown in Table~\ref{table5}, fine-tuning results in an increase of 2.43 percentage points in macro-AP, but no obvious improvement in micro-AP. %We analyze this is because our approach already has a good generalization in the novel set, fine-tuning is not very helpful.

\subsubsection{Word Embeddings}
Table \ref{table3} compares the results we obtained with different word embeddings: FastText \cite{bojanowski-etal-2017-enriching}, Word2Vec \cite{mikolov-etal-2013-linguistic}, GloVe \cite{glove} and BERT \cite{DBLP:conf/naacl/DevlinCLT19}. We also experimented with GloVe vectors of 50, 100 and 200 dimensions obtained from the official GloVe project (Wikipedia 2014 + Gigaword 5)\footnote{https://nlp.stanford.edu/projects/glove/}. To obtain pre-trained label embeddings from BERT, we followed the setting from \cite{aligning}, using BERT-base with masking. %\nb{We will need to clarify which variant we're referring to}\yankun{The current BERT is BERT\_base\_mask\_last. Means we take the vector from the last layer.
In this case, word vectors are obtained by taking the average of 1000 contextualized vectors. As can be seen in Table~\ref{table3}, the best results are obtained for GloVe-300.
%For the word embeddings of GloVe series, we analyze low-dimensional word embedding may be not enough to contain text information to guide the visual model. \textbf{I am not very clear how to explain further, especially for FastText and Skip-Gram.}

\subsection{Qualitative Analysis}
Figure \ref{fig_vis} illustrates which regions are selected by the proposed attention mechanism. For better visualization, we take the values before the softmax layer as the scores of the local features, and we take the sum of attention weights across all attention heads. As the example of the bicycle shows, the model is often successful in identifying the most relevant image region. The second example furthermore shows that the attention weights are indeed label-specific. In this example, the model correctly selects the frisbee or the duck depending on the selected label. This is despite the fact that no images with these labels were present in the training data. On the other hand, as the last example shows, for labels that are semantically closely related, such as \emph{zebra} and \emph{cow} in this case, word vectors are not sufficiently informative. For both labels, the model correctly selects the group of animals, but it fails to make a finer selection.

\section{Conclusion}
We introduced the first metric-based method for multi-label few-shot image classification. The main idea is to use word vectors to obtain noisy prototypes, which are then used to implement an attention mechanism. This attention mechanism aims to construct the final prototypes by aggregating the local features of the support images. Our model achieved substantially better results than existing models, both on COCO and a newly proposed split of PASCAL VOC. Moreover, an important advantage of our model is that it can be used without fine-tuning during the testing phase.
%We are the first work to propose a metric-based method for multi-label few-shot image classification~(ML-FSIC). After training the model in the base set, we can directly predict labels for images  which contains novel categories based on a few-shot labeled examples. We are also the first end-to-end approach in ML-FSIC. We explore a simple yet efficient method to combine word embeddings to solve ML-FSIC problems. We have re-standardized the split the dataset, and proposed a more generative sampling mechanism. More evaluation metrics are introduced to test ML-FSIC model performance more comprehensively. We also proposed a new ML-FSIC benchmark dataset based on VOC, and our approach can achieve the state-of-the-art performance on both COCO and VOC dataset. 

%\begin{figure}[t]
%\centering
%\includegraphics[width=0.9\columnwidth]{figure2} % Reduce the figure size so that it is slightly narrower than the column. Don't use precise values for figure width.This setup will avoid overfull boxes.
%\caption{Using the trim and clip commands produces fragile layers that can result in disasters (like this one from an actual paper) when the color space is corrected or the PDF combined with others for the final proceedings. Crop your figures properly in a graphics program -- not in LaTeX}.
%\label{fig2}
%\end{figure}

\section{Acknowledgments}
This research was supported in part by the National Key R\&D Program of China (2020YFB1805400); National Natural Science Foundation of China (62072010); Capital Health Development Scientific Research Project (Grant 2020-1-4093); Clinical Medicine Plus X - Young Scholars Project, Peking University, the Fundamental Research Funds for the Central Universities; HPC resources from GENCI-IDRIS (Grant 2021-[AD011012273] and ANR CHAIRE IA BE4musIA. Shoaib Jameel is supported by NVIDIA Academic Hardware Grant; 

\bibliography{aaai22.bib}

\ifappendix
\appendix

\section{Additional Ablation Analysis}

\begin{table}[t]
\centering
\footnotesize
\begin{tabular}{ccc}
\toprule
\textbf{Number of Heads} & \textbf{Macro-AP} & \textbf{Micro-AP}\\
\midrule
1          & 30.06 & 20.39 \\
2          & 35.42 & 27.24 \\
4          & 36.68 & 27.02 \\
6          & 35.61 & 27.33 \\
8          & 42.84 & 35.30 \\
10         & 42.00 & 34.35 \\
12         & 42.70 & 35.96 \\
14         & 40.18 & 35.75 \\
16         & 41.53 & 35.66 \\
\bottomrule
%\noalign{\smallskip}
\end{tabular}
\caption{Impact of the number of attention heads.}
\label{table4}
\end{table}

\subsubsection{Number of Attention Heads}
In Table \ref{table4} we analyze the importance of using multiple attention heads. Note that when changing the number of attention heads, we keep the overall dimensionality of the prototypes fixed; e.g.\ when doubling the number of attention heads we halve the dimensionality of the query, key and value vectors. %For example, assume the hidden size is 512, if there is only 1 head, then the dimensionality of $K$, $Q$ and $V$ is the 512; But if there are 2 heads, then the dimensionality of $K$, $Q$ and $V$ is the 256. Thus, we can make sure the whole parameters of different number of head settings are the same. 
The results in Table~\ref{table4} show that with fewer than 8 attention heads, the performance is notably lower, while there is no obvious benefit to increasing the number of attention heads beyond 8.
%with the increasing of the number of attention heads, we can get better results, especially after 6 attention heads. Although the whole parameters are the same no matter how many attention heads are, more attention heads can help model attend more information from different feature subspaces, which can fully explore the relationship between visual features and word embeddings. 

\subsubsection{Scale effects}
As our model uses local features with a fixed dimensionality, there is a risk that some objects are overlooked, when they are too small in the original image. To analyze this effect, we carried out a simple experiment based on ResNet-101, where we increased the image size from $224 \times 224$ to $336 \times 336$ and the corresponding feature map size from $2048 \times 7 \times 7$ to $2048 \times 11 \times 11$. As a result, the micro-AP and macro-AP on COCO increase from 35.30 to 36.41 and from 42.84 to 43.79 respectively. However, it should be noted that this improved performance comes at a higher computational cost.

\subsubsection{Local Features for Query Images}
In our method, we compare the global features of query images with the local features of prototypes. We also experimented with a variant in which the local features of the query image are compared with the local features of the prototypes, where the sum of the top-k values is then used as the similarity score. This variant achieves 35.59 micro-AP and 42.91 macro-AP. This represents a small improvement over our main model (35.30 micro-AP and 42.84 macro-AP), which again comes at the a significantly increased computational cost.
\fi
\end{document}